\documentclass[a4paper,12pt]{article}
\usepackage[dvips]{graphicx}
\usepackage{latexsym}
\usepackage{amsfonts}
\usepackage{amssymb}
\usepackage{amsmath}
\usepackage{amsthm}
\usepackage{pstricks}

\newtheorem{theorem}{Theorem}[section]

\newtheorem{definition}[theorem]{Definition}

\newtheorem{example}[theorem]{Example}

\def\C{ {\mathcal{C} }}

\def\P{\mathbb{P}}

\begin{document}

\title{Prediction with Restricted Resources and Finite Automata}

\author{Finn Macleod, James P Gleeson, MACSI, \\
Department of Mathematics and Statistics, University of Limerick, \\Ireland}

\date{}

\maketitle
\abstract
We obtain an index of the complexity of a random sequence by allowing the role of the measure in classical probability theory to be played by a function we call the generating mechanism. Typically, this generating mechanism will be a finite automata. We generate a set of biased sequences by applying a finite state automata with a specified number, $m$, of states to the set of all binary sequences. Thus we can index the complexity of our random sequence by the number of states of the automata. We detail optimal algorithms to predict sequences generated in this way.


\medskip
{\bf Mathematical Subject Classification:} {\bf ??}

\section{Generating Mechanisms}

We explore a finite setting for the problem of prediction. In particular we are interested in an index of the complexity of a random sequence. In this paper, the role of the measure in classical probability theory will be played by a function we call the generating mechanism. Typically, this generating mechanism will be a finite automata. We generate a set of biased sequences by applying a finite state automata with a specified number, $m$, of states to the set of all binary sequences. Thus we can index the complexity of our random sequence by the number of states of the automata.

We will show the prediction algorithms which minimise average error for varying degrees of knowledge about the generating mechanism. We will then show how the index of complexity used can enable us to consider the batch setting - how best to predict after exposure to a given set of training data. This allows an interpretation of Occam's razor - when and how simpler predictors are better.

Finally we discuss the case of prediction with restricted resources, again utilizing the number of states of the generating mechanism as our index of complexity.

\section{Mathematical Setting.}

\begin{figure}
\centerline{\includegraphics[width = 10cm]{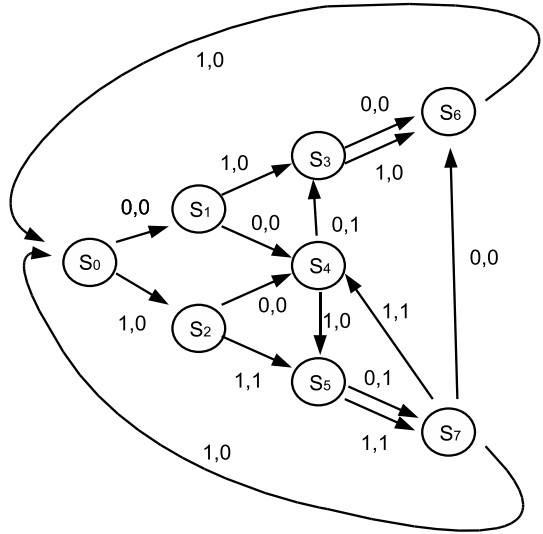}}
\caption[Automata Example] {An example of the type of finite automata known as a Meally Machine with 7 states. The active state is initially $S_0$ It changes according to an input sequence, for example 001111 would cause the following order of states to be active: $S_0S_1S_4S_5S_7S_0S_2$, and the output sequence would be 000100.  }
\label{f:automataexample}
\end{figure}

\begin{figure}
\centerline{\includegraphics[width = 14cm]{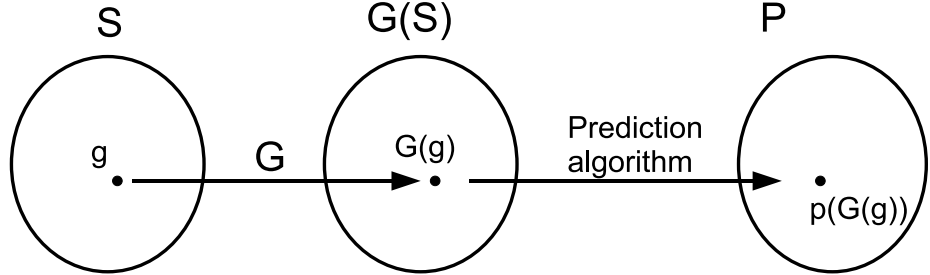}}
\caption[Prediction Setup] {The prediction setup. $S^t$ is the set of binary sequences of length $t$. $G(S^t)$ is the set of possible outputs of the automata $G$. $G(g)$ is a particular element in $G(S^t)$. In a prediction setting, $G(g)$ represents the observed data. We try to predict $G(g)_t$ given $G(g)_1 \ldots G(g)_{t - 1}$ in the best possible way; specifically, we design a prediction algorithm to minimize the error metric. }
\label{f:predictionsetup}
\end{figure}

We consider the set of all length $t$ binary sequences, $S^t = \{0,1\}^t$, which we call the \emph{generating sequences}. We consider them acted upon by a particular finite state automata, $G$, which we will call the \emph{generating mechanism}. We define a finite automata as follows:
\begin{definition}
A finite automata is a system consisting of a set of states S, a transition function $f: S \times \{0,1\} \to S$, and an output function $g:S \times \{0,1\} \to \{0,1\}$, together with an element of S designated as the `active state', initially labelled as $S_0$. Upon receiving a binary input sequence, the active state will change as specified by the transition function, and at each transition will output according to the output function. See fig \ref{f:automataexample}.
\end{definition}
For more on finite automata, see any introductory textbook, eg. \cite{K97}.
\begin{example}
\begin{enumerate}
\item A ring automata that creates a periodic sequence out of any input sequence. $G(S)$ contains only one element.
\item A shift automata that maps all sequences to a shifted version. eg 010111 goes to 0010111. This can be implemented in two states.
\end{enumerate}
\end{example}
This particular finite state automata generates a new set, the set of output sequences, $G(S^t)$. In $G(S^t)$, particular sequences may appear several times, or not at all. We consider possible algorithms, $p(G(S^t))$, of predicting the $i$th element of $G(S^t)$ given all elements up to and including to $i-1$. We answer the following question under certain conditions: ``After observing a sequence $G(g)$ in $G(S^t)$ up to time $t$, what is the best way of predicting the next element in the sequence?". See fig \ref{f:predictionsetup}.

In this article, we define the optimal prediction algorithm, $p$, in several cases, using the average error as the metric of performance (we define the average error below). We calculate the average error associated with these cases as a function of the structure of the generating mechanism(s) involved. Specifically we deal with the following cases in order:
\begin{itemize}
\item We know the structure and active state of $G$ at all times $t$.
\item We know the structure of $G$, but no information as to which state is active
\item We know that $G$ is one finite automata from a known set of finite automata.
\end{itemize}
We then proceed to the case where we have restricted resources. That is, we are predicting a mechanism which could have up to $m$ states, using the automata with $k$ states or less.
How best should we predict the output of such a mechanism, when we don't know the generating sequence?

First we consider metrics for the performance of any prospective predictor.

\section{Measuring the Error}

We consider two measures of the error of an arbitrary prediction algorithm applied to elements of $G(g)$. First define the error:
 \begin{equation}
E^t(g)  = \frac{1}{t}\sum_{i=1}^t G(g)_i \oplus p(G(g))_i,
\end{equation}
where $\oplus$ denotes binary summation modulo 2. Then the average error is
\begin{equation}
E_{ave}:= \frac{1}{2^t}\sum_{g \in S^t} \frac{1}{t} \sum_{i=1}^t G(g)_i \oplus p(G(g))_i,
\end{equation}
and the worst case error is
\begin{equation}
E_{wc}:= \max_{g \in G} E^t(g).
\end{equation}

We could consider other metrics to optimise eg, prediction paths with error above a specified fraction $t$ count are unacceptable, and error below $t$ count as acceptable, find a predictor which maximises total count of acceptable sequences. Here we only consider the average error, $E_{ave}$.

\subsection{Perfect Knowledge - known active state and structure}

\begin{figure}
\centerline{\includegraphics[width = 12cm]{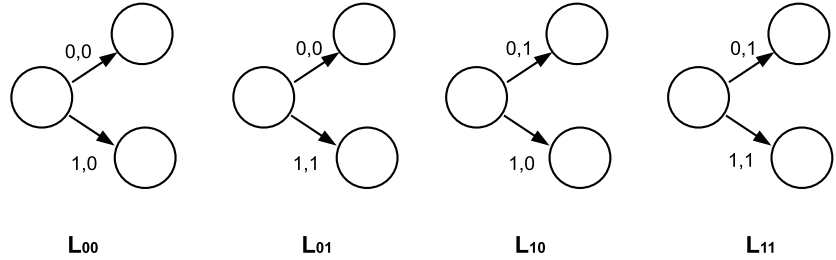}}
\caption[State Types] {States can have one of four different input output combinations. In the figure the transitions are labelled (input, output). If we know the active state of a generating mechanism is of type $L_{00}$ or $L_{11}$ then we can be sure of the next output. In the two other situations (the unbiased states) the output digit will depend on the input. }
\label{f:statetypes}
\end{figure}

Suppose we know the structure and active state of $G$. We are still only able to determine the output digit from the generating mechanism for certain situations. Every active state has a transition from it corresponding to an input of 0, and a transition corresponding to an input of 1. Each transition produces an output of either 0 or 1, and thus we have four possible situations. We label them by their output digits: $L_{00}, L_{01}, L_{10}, L_{11}$. See fig \ref{f:statetypes}.

For situations $L_{00}$, $L_{11}$, whatever the next digit of the generating sequence, we can be sure about the next digit. We call these type of states, with output transitions of case $L_{00}$ and case $L_{11}$, \emph{biased states}. For $L_{01}$ and $L_{10}$, we will be wrong for 1 possible generating sequence digit, and correct for another.

Thus even if we know the active state, and structure of mechanism, the best we can predict depends on the frequency of occurrence of biased states. If the number of times a state $s$ is active over the first $t$ timesteps of $g$ is $a_t(s)$ say, then
\begin{equation}
E_t(g)  = \sum_{\mbox{\{$s$: $s$ is unbiased\}}} a_t(s).
\end{equation}

The frequency of occurrence of a particular state $s \in G$, over the first $t$ digits of a sequence $g \in S$ is defined as
\begin{equation}
f^t(s,g):=  \frac{1}{t} a_t(s).
\end{equation}
We get the average frequency of occurrence by averaging this over the set $S$ and taking the limit:
\begin{equation}
f(s):= \lim_{t \to \infty} \frac{1}{2^t} \sum_{g \in S^t}f^t(s,g).
\end{equation}
Thus if we always know the active state, the average long term error, $E(G)$ will be:
\begin{equation}
E(G)= \sum_{\mbox{\# unbiasedstates}}f(s),
\end{equation}
which is an upper bound on the average error of any prediction algorithm.

\subsubsection{Calculation of state frequencies, $f(s)$, for certain machine structures}

We represent some of the information contained in the transition function of $G$ by an adjacency matrix $A$ - with $A_{ij}$ being the number of possible transitions from state $i$ to state $j$. Thus $A$ contains entries of either 0,1 or 2. We can determine $f(s)$ from knowledge of the adjacency matrix $A$ of the mechanism $G$.

The number of paths leading from state $i$ to state $j$ in $t$ steps is given by $ij$ th entry of the $t$'th power  of the adjacency matrix thus:
\begin{eqnarray}
f(s) &=& \frac{1}{2^t}\sum_{g \in S^t}\frac{1}{t} a_t(s)\\
 &=& \frac{1}{t2^t}\sum_{g \in S^t} a_t(s)\\
 &=& \frac{1}{t2^t}\sum_{i=1}^t A^t_{s_0s}
\end{eqnarray}

Now the adjacency matrix of a mechanism $G$ can be normalised by a factor of $1/2$, and this normalised adjacency matrix has rows which sum to 1. Call the normalised matrix $N$. We thus examine the limit:
\begin{equation}
\lim_{t \to \infty} \frac{1}{t}\sum_{j=1}^t N^i_{s_0s}
\end{equation}

We now borrow a standard result from the theory of Markov Chains (see any introductory text on the subject, eg. \cite{N98}) If $N$ is a irreducible and aperiodic, the limit operation
\begin{equation}
\lim_{t \to \infty} (N^t)_{s_0s}  = \pi_s
\end{equation}
defines a stationary vector, and that this vector is the largest eigenvector of N (with entries summing to 1). One can show that this result implies
\begin{equation}
\lim_{t \to \infty} \frac{1}{t}\sum_{i=1}^t N^i_{s_0s} = \pi_s.
\end{equation}
Thus if $N$ is irreducible and aperiodic, $f(s)$ can be calculated by determining the largest eigenvector of the normalised adjacency matrix of the generating mechanism.

It remains to prove that time averaging allows us to drop the aperiodic condition.
\begin{flushright}
$\blacksquare$
\end{flushright}

\section{A known Mechanism, but with unknown active state}

Suppose we wish to predict an output sequence, $G(g)$ at time $t$, given only the structure of $G$, its initial state and the observed data sequence $G(g)$ up to time $t-1$.

\subsubsection{Optimal prediction}

We now detail the optimal prediction algorithm in this case. First we make the following definition:
\begin{definition}
Given a generating machine $G$, we say a generating sequence $g$ is consistent up to time $t$ with output sequence, $G(g')$, if the first $t$ digits of $G(g)$ agree with $G(g')$. We also say that sequences $g$ and $g'$ are consistent with each other if they are both consistent with the same output sequence. Because the operation of consistency forms an equivalence relation, we can partition the set $S^t$ of generating sequences into sets $C^t_{G(g)}$ defined by the output sequence $G(g)$. We call these sets the \textit{consistency classes} --- each sequence in a consistency class is consistent with all other sequences in that class.
\end{definition}
Now, we can write the average error:
\begin{equation}
\sum_{g \in S^t} G(g)_{t+1} \oplus p(G(g))_{t+1}
\end{equation}
as a sum over the consistency classes
\begin{equation}
\sum_{C \in \C} \sum_{g \in C} G(g)_{t+1} \oplus p(G(g))_{t+1}.
\end{equation}
We note that because the observed data, $G(g)$ is the same for all $g$ in a consistency class, the prediction $p$ will be identical for all elements in the class. If we desire to choose our predictor, $p$, in order minimize the average error, then for each class, $p(G(g)_1\ldots G(g)_t)$ should be 0 if $G(g)_{t+1}=0$ more often than $G(g)_{t+1} = 0$. Vice versa, if $G(g)_{t+1}=1$ more often than $G(g)_{t+1} = 0$, then $p(G(g)_1\ldots G(g)_t)$ should be 1.

More precisely, let the number of $g$ for which $G(g)_{t+1}=0$ be $\#p_{t+1}$. Let the number of $g$ for which $G(g)_{t+1}=1$ be $\#q_{t+1}$. We can determine these quantities from the knowledge of the location of the active states for each generating sequence within a consistency class. Then:
\begin{eqnarray}
\#p_{t+1} = |L_{00}| + |L_{01}| + |L_{10}|\\
\#q_{t+1} = |L_{11}| + |L_{01}| + |L_{10}|
\end{eqnarray}

Now the combined error
\begin{align*}
\mathop{\sum_{g \in C^{t+1}_{G(g)}}}_{G(g)_{t+1}=0} G(g)_{t+1} \oplus p(G(g))_{t+1} + \mathop{\sum_{g \in C^{t+1}_{G(g)}}}_{G(g)_{t+1}=1} G(g)_{t+1} \oplus p(G(g))_{t+1}\\
= \mathop{\sum_{g \in C^{t+1}_{G(g)}}}_{G(g)_{t+1}=0} 0 \oplus p(G(g))_{t+1} + \mathop{\sum_{g \in C^{t+1}_{G(g)}}}_{G(g)_{t+1}=1} 1 \oplus p(G(g))_{t+1}\\
= \left\{
\begin{array}{ll}
|\{C^{t+1}_{G(g)}: G(g)_{t+1}=1\}| & \mbox{if $p(G(g))_{t+1}=0$}\\
|\{C^{t+1}_{G(g)}: G(g)_{t+1}=0\}| & \mbox{if $p(G(g))_{t+1}=1$,}
\end{array}
\right.
\end{align*}
Thus to minimize the error, we define:
\begin{equation}
p(G(g)_1\ldots G(g)_t):= \left\{
\begin{array}{ll}
0 & \mbox{if $\#p_{t+1}\geq \#q_{t+1}$,}\\
1 & \mbox{if $\#p_{t+1} < \#q_{t+1}$,}
\end{array}
\right.
\end{equation}
Then

\begin{align*}
\mathop{\sum_{g \in C^{t+1}_{G(g)}}}_{G(g)_{t+1}=0} G(g)_{t+1} \oplus p(G(g))_{t+1} + \mathop{\sum_{g \in C^{t+1}_{G(g)}}}_{G(g)_{t+1}=1} G(g)_{t+1} \oplus p(G(g))_{t+1} \\
 = \min\{\#p_{t+1}, \#q_{t+1}\}\\
\end{align*}

\subsubsection{Average Error}

Given the structure of $G$, can we determine the average error in a similar fashion to the case where we always knew the active state?

\begin{equation}
E(G) := \lim_{t \to \infty} \frac{1}{2^t t}\sum_{g \in S^t}\sum_{i=1}^t G(g)_i \oplus p(g)_i.
\end{equation}

To calculate the best prediction we note we only require the knowledge of the following:
\begin{definition}
The consistency vector of an output sequence $G(g)$ is a size $k$ vector, where the $i$'th entry contains the number of generating sequences $g$ active at state $i$ which are consistent with $G(g)$.
\end{definition}
From the structure of $G$ we can define two matrices $B,C$ which evolve the consistency vector under the inputs of 0 and 1 respectively. We note that $B+C = A$.

We note that one can reach the same ratio of active states (and thus make the same prediction) by more than one generating sequence. Our predictions and errors are only determined by the ratio of generating sequences active at states $s_1$ to $s_k$. Thus we can be somewhat more accurate with our choice of equivalence classes. We can define a space of all possible ratios, \textit{ratio space}. If we know the time average of the number of sequences active at each point in ratio space, then we can calculate the average error.

We have not yet determined this time average, and thus determining a closed form for the limit of $E_{ave}$ as $t$ tends to infinity, in terms of the adjacency matrix of the generating mechanism, is a problem that remains to be solved.

\section{Selecting from a set of automata}

We now consider the setting where we do not know the particular generating mechanism, we only know that it is a member of a prescribed finite set of generating mechanisms $\{G_1, \ldots G_n\}$. In this case, the best prediction algorithm we can use is the same as the previous case of a known mechanism but unknown active state. We replace the tracking of all possible consistent generating sequences with the tracking of all consistent $(G_i,g)$ pairs. At a given timestep, we make our prediction by comparing the number of $(G_i,g)$ pairs which predict 0 with the number that predict 1. We predict a 0 if the number predicting 0 is larger than the number predicting 1, and we predict 0 otherwise. The error associated with such a prediction will be the minimum of these numbers.

We conjecture that the asymptotic error $E_{ave}$ of this situation will be the same as the previous case. Secondly, this may end up having significant computational cost. If there are symmetries in the set of $\{G_1,\ldots G_n\}$, then we may be able to increase the speed of this exhaustive search algorithm significantly (and possibly perform nearly as well).

\section{The batch setting - Occam's Razor on a finite set of mechanisms}

We now consider a batch setting - that is, given the performance of different predictors over a training data set, how should one choose the predictor with the best performance on future data? Even if we have a predictor which makes no error over the data set, this does not guarantee anything about the performance of that same predictor over future data. We set up the problem precisely as follows:

\textit{
We have an unknown generating mechanism $G$ in some finite set of mechanisms, $\{G_1\ldots G_n\}$. $G_m$ produces a particular output sequence for a given generating sequence $g$, which we call the training data: $o_1\ldots o_t$. We have a set of predictors $\P = \{P_i\}$. Given the number of errors each predictor, $P_i$, makes on the training data, how does one pick the predictor with minimum error on continuations of the data: the sequence $o_{t+1}\ldots o_{T}$?}

The first step is to collect information about the `likelihood' for each generating mechanism that might have been responsible for the training data. We represent this as a set of generating sequence, generating mechanism pairs, $(g,G)$, whose output results in the training data:  $G(g) = o_1 \dots o_t$.

Any generating sequence could be responsible for continuation of the data. However, we know that only a certain set of generating sequence, generating mechanism pairs could have resulted in the training data. For each predictor, we can calculate the asymptotic performance of the predictor applied to a particular generating mechanism.  We make the following definition:
\begin{definition}
The average error of a predictor $P$ with respect to $G$ at time $t$, is the average number of errors made by $P$ when trying to predict an output sequence $G(g)$, and then averaged over all generating sequences $g$ of length $t$.
\begin{equation}
E^t(P, G):= \frac{1}{t 2^t} \sum_{g \in S^t} \sum_{i=1}^{t} P(G(g)_1\ldots G(g)_{i-1}) \oplus G(g)_i.
\end{equation}
\end{definition}
We can determine this quantity for each $P$ by an exhaustive search over all $g$.

We consider the average performance of a predictor $P_k$ over this set of pairs.
\begin{equation}
\label{e:1}
\sum_{g, G_m} E^{T-t}(P_k, (g, G_m)),
\end{equation}
where we include the generating sequence in the definition of the predictor, because at time $t$ it defines the starting state of $G_m$.

Finding the $P_k$ which minimizes quantity (\ref{e:1}) gives the best predictor to use. This predictor \textit{may not be the one with the best performance over the training data}. One can observe that we do not use the number of errors that each predictor makes on the training data in the calculation directly.

Again there are opportunities for implementing these algorithms more efficiently by using symmetries.

We can also calculate this quantity for types of predictors other than automata, for example decision trees.

\section{Restricted numbers of states}
In the above setting, we have assumed that whilst we have a restricted number of predictors, we have an infinite amount of resources to allow us to make the best choice. We now consider the problem where the resources with which we implement \textit{and} select the model are restricted. That is, we have X memory states for both determining the best predicting algorithm and implementing it.

If one takes the resources restriction to be represented by the number of states in an automata, then we have set this problem up as one of finding the `best' predicting automata with a number of states.

Now, we have to specify our method of selecting the best automata, \textit{before} we see the training data. That is we must choose our \textit{automata} before we see the training data. The individual predictors are encapsulated in the structure of the single automata chosen as our best method. After moving around according to the training data, these states must then perform well on the actual data.

We can conduct an exhaustive search to find these optimal automata for finite values of $t$.

\section{Comments}

Utilising the number of states of a finite automata as an index of the complexity of a random sequence allows one to ask quantitative questions about prediction with restricted resources. Other indices are certainly possible.

We are also interested in understanding how well one can predict a k state automata with an $m<k$ automata. Related work has been done - see for example, Meron and Feder's paper ``Finite-Memory universal prediction for individual sequences" \cite{MF04}. We would like to see this extended to a formula describing how well one can predict an unknown automata of size $k$ with automata of size $m<k$.

We note that numerical application of these algorithms is computationally intensive. For example the number of binary automata grows quickly with the number of states, $k$. eg. $(2k)^{2k}/k!$, see \cite{HP67},\cite{H64}. Or  see \cite{R65} for enumeration of strongly connected automata (any state is accessible from any other state). Speeding up these kind of exhaustive searches is of great interest. We note Helmbold and Schapire's work \cite{HS95} in efficient approximation of a prediction algorithm using the symmetries of underlying predictors. We speculate that a similar result may be applicable to finite automata.

This research was made possible by funding from Science Foundation Ireland through MACSI, and programme 06/IN.1/I366. We would like to thank V. Vovk for his comments.

\bibliographystyle{beta}

\end{document}